*Review Article*

# Water-Based Metaheuristics: How Water Dynamics Can Help Us to Solve NP-Hard Problems

**Fernando Rubio and Ismael Rodríguez**

*Departamento de Sistemas Informáticos y Computación. Facultad de Informática, Universidad Complutense de Madrid, 28040 Madrid, Spain*

Correspondence should be addressed to Fernando Rubio; fernando@sip.ucm.es

Received 9 January 2019; Revised 13 March 2019; Accepted 18 March 2019; Published 2 April 2019

Academic Editor: Alex Alexandridis

Many water-based optimization metaheuristics have been introduced during the last decade, both for combinatorial and for continuous optimization. Despite the strong similarities of these methods in terms of their underlying natural metaphors (most of them emulate, in some way or another, how drops collaboratively form paths down to the sea), in general the resulting algorithms are quite different in terms of their searching approach or their solution construction approach. For instance, each entity may represent a solution by itself or, alternatively, entities may construct solutions by modifying the landscape while moving. A researcher or practitioner could assume that the degree of similarity between two water-based metaheuristics heavily depends on the similarity of the natural water mechanics they emulate, but this is not the case. In order to bring some clarity to this mosaic of apparently related metaheuristics, in this paper we introduce them, explain their mechanics, and highlight their differences.

## 1. Introduction

It is common for an engineer or scientist to eventually face an NP-hard optimization problem during her career. This is an unfortunate encounter indeed: even if the problem is also in NP (so the quality of potential solutions can be computed in polynomial time), optimal solutions cannot be found in polynomial time unless P = NP. Yet approximate solutions must be found somehow. For some problems we can apply problem-specific polynomial-time heuristics guaranteeing some performance ratio (i.e., ratio between found solutions and optimal solutions) in the worst case, in particular for problems in APX (or, better, in PTAS or FPTAS). (The interested reader is referred to [1] for a detailed description of approximation classes.) Unfortunately, for other optimization problems (in particular, those that are Log-APX-hard or worse) this is not possible, so algorithms providing good solutions on average (but not guaranteeing any performance ratio in the worst case) are needed.

Nature has been broadly used as a source of inspiration for developing heuristic methods of the latter kind, in particular for constructing general-purpose metaheuristics that can be adapted to many particular problems. These metaheuristics include Genetic Algorithms (GA) [2], Ant Colony Optimization (ACO) [3, 4], and Particle Swarm Optimization (PSO) [5]. It is worth noting that the natural processes these methods copy do not aim at solving anything that can be formally stated as an actual computational NP-hard *optimization problem* (that is, maximizing or minimizing a given known and fixed numerical function). For instance, regarding the natural process GA are inspired on, surviving other species in a real environment is not a well-defined optimization computational problem because the function to be optimized is not known and it even changes over time (other species constantly evolve in unknown ways). Concerning ACO, the problem of finding the shortest path between two nodes in a graph is not NP-hard in its most basic form, as it is solved in polynomial time (e.g., by Dijkstra's algorithm [6]). As for PSO, finding a food source in an environment with no formal definition of where the food is or can be (i.e., with no function to be maximized) is not an optimization problem either. In fact, we do not know yet how nature would solve an NP-hard optimization problem (in any way other than ours) if it ever had that purpose, as we seem to be the only beings facing that particular goal. Anyway, the similarities between natural goals and problem optimization goals have yielded good heuristics for optimization problems, but not just due to the metaphors they are based on, but rather due to their proper translation and adaptation into suitable algorithms.

A little more than a decade ago, researchers also found inspiration on another natural source: the water dynamics. The way water drops join and collaboratively form paths (flows) of water toward their destination (e.g., the sea) in real environments has been seen as a source of inspiration, similarly to the way chromosomes, birds, or ants interact with each other according to some simple rules to transform themselves into better solutions (in the case of chromosomes and birds) or collaboratively draw a solution on a kind of solution canvas by iteratively modifying it (in the case of ants). Starting from the basic idea of emulating how drops form flows, the research community has independently developed many optimization algorithms (see, e.g., [7–10]) which, despite the common original metaphor they are based on, are quite different (sometimes radically different) in terms of what really matters: the mechanics of the algorithm and the properties of the search this algorithm induces. The actual interest of any new metaheuristic lies in the computational steps it follows and the kind of searching properties these steps unleash (compared to previous methods in the literature), regardless of the beauty of the metaphor it is based on—which, in fact, is unnecessary to develop a new good algorithm. Despite this, for some reason the community expects different algorithms based on the same natural process to expose similar properties. In fact, in the last years the metaheuristics community has given excessive importance to the metaphors the new methods are based on, to the extent that the area has witnessed an explosion of new methods whose novelty is more based on the alleged novelty of the natural metaphor they are based on rather than on the novelty of the resulting algorithm [11].

In order to clarify this research domain and avoid any confusion between metaphors and algorithm mechanics, in this paper we introduce all the optimization algorithms based on water dynamics we have found in the literature to date, we explain the particular aspects of the water dynamics each method is based on, and we describe the resulting algorithm in order to understand the actual optimization mechanism it poses and its important differences from other methods based on the same basic natural metaphor. We hope this paper will bring some clarity to an area full of unfortunate notation and metaphorical coincidences. It is worth noting that this paper will focus on the application of water dynamics to the design of optimization algorithms, so we will not consider its usage in other computational domains. For instance, the water metaphor has been used for decades in the field of image processing by means of concepts like *watersheds* (see, e.g., [12, 13]), which have been used, for instance, to analyze borders or to obtain binarizations of an image to simplify character recognition (see, e.g., [14]).

The rest of the paper is structured as follows. In the next section we describe the water-based metaheuristics that have been designed to deal with *combinatorial* optimization problems, whereas Section 3 presents water-based metaheuristics dealing with *continuous* domain optimization problems. Finally, in Section 4 we present our conclusions.

## 2. Metaheuristics for Combinatorial Optimization Problems

Three different water-based metaheuristics for combinatorial optimization appeared nearly at the same time during 2007. However, although all of them were directly inspired by observing how water flows from high to low altitudes, the concrete algorithms are completely different. In the rest of this section we describe these metaheuristics and also a fourth method that appeared ten years later, which is closely related to one of the three initial methods. For each metaheuristic, we describe how it works and we also comment on its main application areas.

*2.1. RFD: River Formation Dynamics.* River Formation Dynamics (RFD) was the first water-based metaheuristic published. It was published in August 2007 in [7], only one month before the publication of Intelligent Water Drops (IWD) in [8].

River Formation Dynamics simulates how water drops collaboratively form rivers in their way to the sea. Drops tend to move through steeper down slopes with higher probability, and they extract soil from the ground while falling through steep slopes. This sediment is deposited later in flatter areas or in lakes. This way, the altitudes of locations around the environment iteratively change. In algorithmic terms, RFD entities (drops) iteratively modify the values attached to the locations of a graph while traversing it, like ants do in ACO. The paths of decreasing altitudes formed on the graph during this process will eventually constitute the problem solutions. The values attached to graph locations, used in RFD to represent *altitudes*, are attached to nodes (contrarily to pheromone trails in ACO, which are attached to edges). Actually, RFD can be seen as a *gradient-oriented* version of ACO, because the probabilistic choice of where entities move next is not proportional to the values attached to available edges (the amounts of pheromone trail in ACO), but to the *difference* of values between each possible destination node and the origin node (*difference* of altitudes in RFD). This basic but essential difference from ACO makes RFD fulfill new interesting properties, such as trivially avoiding round-trips (as they would imply an ever decreasing cycle, which is impossible), quickly reinforcing newly discovered shortcuts (as they have the same altitude difference as the former path but a shorter distance, which implies steeper slopes on average through the shortcut even *before* any subsequent reinforcement), and quickly eliminating blind alleys (as sediments are accumulated in lakes, eventually turning attractive down slopes into uninteresting flat slopes). On the other hand, the transposition of path subsequences within a formed solution is harder in RFD than in ACO and for the same reasons: decreasing altitudes tend to constitute a natural traversing order through paths in RFD, which does not happen in ACO. The pseudocode of RFD can be summarized as follows:

```
initialize drops
initialize nodes
repeat
  move drops
  erode paths
  deposit sediments
  analyze paths
until stopping criteria met
```

RFD has been used by different research groups to solve a variety of problems. In addition to well-known academic NP-hard problems like the traveling salesman or Steiner tree problems [19, 20], it has also been used in more specific industrial domains, like monitoring electrical power systems [21] or designing VLSI circuits [22, 23]. In particular, due to its intrinsic absence of cycles, it has been used to design several routing protocols in computer networks (see, e.g., [24, 25]), to test systems [26], or to improve robot navigation [27]. A survey describing the main application domains of RFD can be found in [28].

*2.2. IWD: Intelligent Water Drops.* The first paper describing the Intelligent Water Drops (IWD) metaheuristic [8] was published only one month later than the publication of RFD. This method is also a modification of ACO to simulate how water flows from high altitudes to the sea. However, it is clear that IWD and RFD were developed completely independently, as the formalization of IWD is totally different. In IWD, drops also collaborate to form solutions on a graph by modifying some values attached to its locations, like in ACO or RFD. However, IWD does not associate altitudes with nodes like RFD does. In this sense IWD is closer to ACO, as values are associated with *edges* of the graph (*soil* amounts in IWD, like pheromone trails in ACO), and the probabilistic movements of entities are governed by the values attached to possible edges (like ACO) rather than by the *difference* of these values between each possible destination node and the departure node (like RFD). Thus, the differences between the properties of RFD and ACO due to the gradient orientation of the former, mentioned in the previous section, are also differences between RFD and IWD. For instance, in IWD drops need to record the path they have traversed (as in ACO and opposed to RFD because it is not needed to avoid cycles). The main difference of IWD with respect to ACO is that in IWD drops have a *velocity* and an amount of transported *soil*. By using these data, drops can behave differently from ants in ACO. In particular, the velocity is used to decide how much erosion is to be done.

The algorithm works as follows. At the beginning, it is common but not mandatory to create exactly the same number of drops as nodes in the graph. If both numbers are not equal, drops are randomly distributed among the nodes. All drops are initialized with a fixed velocity, zero amount of soil, and a singleton list of visited nodes (including only the node they have been assigned to). Analogously, all edges of the graph are initialized with a fixed amount of soil. In each iteration, each drop moves as follows. As long as the path traversed by the drop so far constitutes a part of a possible solution, it moves to some other node. As mentioned before, the probability of selecting each neighbor node (not visited before) depends on the amount of soil of the corresponding edge: the lower the soil, the higher the probability. After moving the drop, the node is added to the list of visited nodes of the drop, and the drop velocity is updated: the lower the soil in the edge, the larger the velocity increment. Besides, the soil carried by the drop is also updated, adding the same amount of soil that has been taken from the corresponding edge. The process is repeated until a complete solution is found by each drop. After each iteration, the best drop of the iteration is selected. Then, the amount of soil of all edges traversed by this drop through its path (solution) is reduced. This reduction is based on the quality of the solution found by the drop. The pseudocode of IWD can be summarized as follows:

```
initialization
repeat
  create and distribute IWDs
  update visited lists of IWDs
  for each IWD
    choose next path
    update velocity
    compute soil
    remove soil from path
    add soil to IWD
  find iteration-best solution
  update path of iter-best solution
  update total-best solution
until stopping criteria met
```

IWD has also been used by many research groups (some have proposed alternative versions of IWD; see, e.g., [29]) and has been applied to many problems, including classic academic problems such as the traveling salesman problem, the n-queens problem, or the multiple knapsack problem [30], but also other industrial problems. In particular, it has been applied to many scheduling problems (see, e.g., [31–34]), as well as to minimize the energy consumption in a power system that requires meeting a given total power demand at each moment [35]. It has also been used to reduce the energy consumption in wireless sensor networks [36], to deal with the robot path planning [37], and even to compress images [38].

*2.3. WFA: Water Flow-Like Algorithm.* The third and last water-based metaheuristic published in 2007 was the Water Flow Algorithm (WFA) [9]. It also takes inspiration from how water moves from high positions down to the sea following the steepest slopes, but the resulting metaheuristic is not related to ACO—the most related method is the tabu search [39]. Entities do not collaboratively draw solutions on a common graph by modifying the values attached to its locations, but each entity is a candidate solution of the problem by itself. Candidate solutions can move to explore nearby positions, representing other candidate solutions in the search space. The *altitude* of each point in the search space represents the fitness of the corresponding solution. Thus, the lower the value, the better the solution, so that water flowing toward lower positions means improving the quality of its solution.

Initially, a single water flow is created with an empty tabu list, as well as with a given initial mass and velocity so that the momentum of the fluid can be computed. Then, the water flows to lower altitudes. There are four main operations in WFA: splitting and moving, merging, evaporation, and precipitation. In the first operation, the velocity is used to decide whether the flow can move or must stagnate. In case it can move, the available neighbor locations are evaluated and ranked, but those locations appearing in the tabu list of the flow are removed. The momentum of the water flow is used to decide whether the flow should be split into several independent flows or not. In case it is large enough, it is split into two or three flows (depending on the concrete momentum value), while otherwise the flow is not divided. Then, the subflows (or the undivided flow) move to the best ranked neighbors. The mass of the flow is divided into the subflows proportionally to the quality of the corresponding neighbors, so that the most promising positions can be explored more exhaustively, as the corresponding momentum is bigger. Anyway, the total momentum is kept. The tabu lists of the new subflows are equal to the tabu list of the original flow after adding the position that was explored in the previous step. Finally, a special case occurs when all the neighbors are higher than the current position. In such situation, if the kinetic energy of the flow is larger than the potential energy needed to jump to the neighbor position, then the flow can jump. Otherwise, it stagnates.

After the splitting and moving operation, whether two (or more) flows have arrived at the same position is checked. In this case, both flows are merged into a single flow, keeping the overall momentum. Notice that one of the merged flows could be a stagnating flow. In this case, merging it with another active flow can allow it to move again. Regarding the tabu list of the new merged flow, it combines the tabu lists of the original flows.

In each iteration, a part of all of the flows is removed right after the merging operation. By doing so, the water evaporation to the air is simulated. However, the precipitation operation does not occur in all iterations, but only once every $t$ iterations. During the precipitation, a new population of flows is added to the system. As it can be expected, the total mass of the new flows equals the total mass of water that evaporated during the last $t$ iterations. Regarding the positions where it rains, they are obtained by applying random distributions around the current flows. The initial velocity of the new flows is the same as the initial velocity of the first created flow, while the tabu list is copied from the tabu list of the closest water flow.

In addition to regular precipitation, *enforced* precipitation takes place in case all flows stagnate. In this case, the method does not need to wait until all of them get evaporated after several iterations. Instead of that, they are forced to evaporate in a single step, and raining takes place afterwards.

As it can be observed, the size of the population varies along the execution of WFA. It starts with a single entity, next the splitting operation increases the population (allowing the exploration of more paths), and merging can reduce the population again, giving more energy to those positions that have been found through different paths. The concept of using a varying number of entities could be easily exported to other population-based metaheuristics. The pseudocode of WFA can be summarized as follows:

```
initialization
repeat
  flow splitting and moving
  flow merging
  water evaporation
  if rainfall required then
    precipitation
    flow merging
  if new best solution found then
    update best solution
until stopping criteria met
```

WFA has also been used by several research groups. It has been used to tackle problems such as the bin packing problem [9], the traveling salesman problem [40], the flow shop scheduling [41, 42], the cell formation problem [43, 44], the quadratic assignment problem [45], and the design of logistic reverse networks [46]. Moreover, it has also been hybridized with fuzzy logic [47].

*2.4. HCA: Hydrological Cycle Algorithm.* The Hydrological Cycle Algorithm (HCA) [48] was introduced in 2017. It takes IWD as a starting point, although four stages of the water cycle are considered: flow, evaporation, condensation, and precipitation. The flow stage, where drops collaborate to modify a common graph, is very similar to IWD. Its main difference is the introduction of the concept of *temperature*, needed to decide when to apply the evaporation stage. The temperature is computed by calculating how much improvement has been obtained during the last iterations of the flow stage. The lower the improvement, the higher the temperature increase. When the flowing stage is improving, the algorithm does not apply evaporation, but when it does not improve enough, it is assumed that the stage has converged to a (probably local) optimum. Hence, the evaporation takes place as a way to restart the system to avoid local optima.

During the evaporation stage, a given proportion of drops is evaporated. The specific drops that evaporate are selected by using a roulette-wheel selection, so that the best drops (i.e., those finding the best solutions) are more likely to evaporate. After the evaporation, the temperature is reduced again, and the condensation takes place. During the condensation stage, drops are together in the clouds, so they can exchange information between them directly. Notice that during the flow stage (or in IWD) drops only affect each other in an indirect way, by modifying a common graph. However, in the condensation stage of HCA, any method can be used to exchange information—actually, the condensation stage is problem-dependent. For each concrete application, a different way to recombine drops information can be used, taking advantage of our information about the specific application domain. After the condensation stage, a new set of drops is obtained from the evaporated drops.

In the last stage, the precipitation takes place. In this phase, the termination condition is evaluated. If it is not met then the precipitation restarts the hydrological cycle. That is, the velocity of water drops is restarted and also the amount of soil of each edge, etc. However, this restart is not done from scratch, as some information from the previous iteration is reused. In particular, the edges that belong to the best drop reduce their amount of soil. The aim of this modification is to emphasize the future search near the best solution found so far. This can be seen as a reinforcement stage.

Let us remark that although HCA uses IWD in the flow stage (the stage where drops modify the common graph to find a solution), the overall method is partially independent of IWD. In fact, IWD could be easily replaced by other metaheuristics like ACO or RFD. The main modification needed to perform such a replacement would be to include a proper definition of temperature in those methods. It would be interesting to analyze in what contexts such modifications could be beneficial for the overall performance.

Let us also remark that the condensation stage is also a completely independent stage, where the aim is to allow a direct communication among drops. In this sense, it would be interesting to analyze if a different metaheuristic could be used in this stage. In this case, HCA could be seen as a framework to combine direct-communication and indirect-communication metaheuristics. The pseudocode of HCA can be summarized as follows:

```
initialization
repeat
  repeat
    for each drop
      choose next node
      update velocity
      update soil and depth
      update carrying soil
    compute solution fitness
    update local optimal
    update temperature
  until temperature == evaporationTemp
  evaporationStage
  condensationStage
  precipitationStage
until stopping criteria met
```

HCA was introduced very recently in [48] and, to the best of our knowledge, it has only been used by its own authors so far. They have used it to solve the traveling salesman problem [49] and also to solve the capacitated vehicle routing problem [50].

*2.5. Comparison.* Given the essential differences among the aforementioned methods in terms of both solution representation and the evolution of that representation, next we analyze the differences among these methods in qualitative terms, focusing on differentiating their internal mechanics and how these mechanics affect their suitability for each kind of problem. A detailed quantitative empirical comparison among the aforementioned metaheuristics is out of the scope of this paper.

Note that the method that differs the most from the others is WFA. Let us remark that in IWD and RFD entities collaboratively work by modifying a common canvas where solutions are constructed, whereas this is not the case in WFA, where each entity represents a possible solution by itself. Thus, the user of WFA has to codify the solutions of the corresponding problem in a very different way from IWD or RFD users.

Regarding the comparison between RFD and IWD, even though they both work by modifying that common canvas (a graph), the codifications users have to perform are relatively different. The main difference between both methods is that IWD is edge-oriented while RFD is node-oriented. That is, in IWD drops modify the attributes of the edges of the graph (soil amounts), whereas in RFD drops modify the attributes of the nodes of the common graph (altitudes). Thus, IWD is more closely related to ACO, whereas RFD can be viewed as a gradient-oriented version of ACO. This relevant difference makes RFD better for dealing with problems where cycles are to be avoided, due to its natural, implicit, and memory-free cycle avoidance. For instance, RFD is very convenient for dealing with different types of covering trees (see, e.g., [51]), but it is harder for RFD to deal with other issues like exchanging subsequences of solutions in problems like the traveling salesman problem.

Regarding the most recent method, note that the core of HCA is IWD. Thus, it inherits most of its characteristics. However, the additional stages introduced in HCA are interesting to allow hybridizing IWD with other techniques. In this sense, HCA seems to be a good solution when IWD stagnates in suboptimal solutions, so that the extra stages can help in exploring other areas out of the watershed of the local minima.

## 3. Metaheuristics for Continuous Optimization Problems

The first water-based metaheuristic devoted to continuous optimization appeared five years later than those for combinatorial optimization. However, during the last seven years, new water-based metaheuristics have appeared every year. In the rest of this section we describe all of them in chronological order, and we also comment on some concrete problems that have been solved by applying them.

*3.1. Continuous Versions of IWD and HCA.* Two of the metaheuristics presented in the previous section (IWD and HCA) have also been used to deal with continuous optimization problems. In fact, the method used in both cases is basically the same. In [52] the authors propose a way to convert a continuous search problem into a discrete environment. The basic idea is the following. Assuming the function to be optimized is $f : R^n \longrightarrow R$ and assuming that we are using $p$ bits to represent each real number, a directed graph with $p * n + 1$ nodes and $2 * p * n$ edges is created. All the nodes are structured in a line, so that there are exactly two directed edges from each node to the next node. One of the edges is labeled 1 and the other edge is labeled 0. By doing so, any path from the first node to the last node represents a binary codification of $n$ real numbers, using $p$ bits for each of them. That is, each path represents a possible point in the search domain, and we can assign the value of $f$ in that point to the path itself. Thus, minimizing (or maximizing) the original function $f : R^n \longrightarrow R$ requires finding the optimal path in this graph.

Once a graph has been obtained from the original fitness function, IWD is used basically as described in Section 2.2. The main modification is that a mutation operator is used after each iteration. This mutation consists in selecting randomly an edge of the solution and replacing it by using the alternative edge. That is, if it was a 1 edge then it is changed by the corresponding 0 edge, and vice versa.

A very similar approach is used in [48] to use HCA in continuous domains, but in this case the representation of the real numbers is done with decimal numbers instead of using a binary representation. Thus, $p*n+1$ nodes and $10*p*n$ edges are used, $n$ being the number of dimensions of the problem and $p$ the number of digits of the decimal representation used for real numbers. Obviously, the 10 directed edges connecting each node with the following node are labeled with digits from 0 to 9.

Let us remark that this method could also be used with any other graph-based metaheuristic, like RFD. The main drawback of this method is that the size of the graph is usually too big when dealing with either large dimensions or high precision. For instance, all the running examples presented in [48] have 6 or less dimensions.

*3.2. WCA: Water Cycle Algorithm.* The Water Cycle Algorithm (WCA) was introduced in 2012 in [53] to deal with constrained continuous problems. It is a population-based metaheuristic where individuals are distributed randomly along the search space during the initialization phase, and then they are divided into three categories, depending on their quality according to the fitness function. The most promising one is called the *sea*. Then, the following $r$ best individuals are considered *rivers*, while the rest are considered *streams*. Inspired by how rivers flow in nature toward the sea, as well as how streams flow toward rivers, in WCA the positions of rivers are modified in each iteration in the direction to the sea, while the positions of streams are modified in the direction to their corresponding river. Each stream is associated with a specific river, and the number of streams associated with each river depends on the quality of the solution found by the river: the most promising rivers have more streams flowing toward them. By doing so, the algorithm explores deeper the most promising regions without abandoning the exploration of other areas.

After each iteration of the algorithm, the fitness of each individual is recomputed in its new position. In case the quality of a stream outperforms the quality of its corresponding river, they swap their roles: the stream becomes the river, whereas the river becomes a stream flowing toward the new river. Analogously, if the quality of a river outperforms the quality of the sea then they exchange roles.

When the position of a river is very close to the sea (or a stream is very close to its river), the evaporation takes place. When the distance is lower than a given parameter $d_{max}$, the river (or stream) is removed. Then, the raining process takes place, and a new stream is created. When a normal stream evaporates, the raining process takes place randomly in the search space (in order to widen the exploration of the search space), whereas when a stream evaporates near the sea, the raining process takes place following a Gaussian distribution around the sea (in order to intensify the exploitation near the current optimum). Finally, the evaporation phase is slightly modified to increase the evaporation probability of the less promising rivers and streams.

The most interesting idea of WCA is probably the hierarchical structure of the population, as opposed to other more classic metaheuristics like Particle Swarm Optimization (PSO) [5], Artificial Bee Colony (ABC) [54], etc. In fact, this idea could be easily adapted to deal with other population-based algorithms, including water-based metaheuristics. It would be interesting to study whether this hierarchicalization can help in improving other metaheuristics, like those described in the next sections. The pseudocode of WCA can be summarized as follows:

```
initialization
calculate cost of each raindrop
determine flow intensity
repeat
  streams flow to rivers
  rivers flow to sea
  for each stream
    if solution better than river then
      exchange stream-river roles
  for each river
    if solution better than sea then
      exchange river-sea roles
  if evaporation condition then
    create clouds and starts raining
  update search intensity parameter
until stopping criteria met
```

WCA has been used in many different applications by different researchers. The original proposal (see [15, 53]), which has been referenced more than 300 times, introduced examples of typical unimodal and multimodal functions, including also typical constrained problems. Later on, it was extended to deal with multiobjective optimization [55]. Regarding industrial applications, it has been applied to optimize the operation of a dam [56], as well as to deal with different issues in power systems, including how to control oscillations by using power system stabilizers [57], how to deal with the economic load dispatch problem in power systems [58], and also how to minimize the probability of power supply loss [59]. Other examples include optimizing the flow of material in supply chains [60] and the control of traffic lights to minimize the total delay of cars [61].

*3.3. SRA: Simulated Raindrop Algorithm.* SRA was introduced in 2014 in [16]. In this case there is not any analogy with the water cycle or even with how rivers flow toward positions with a lower altitude. In fact, the analogy (and also the algorithm) is much simpler. It is based on the splash that takes place when a raindrop hits the ground. In this case, after hitting the ground, part of the water can reach nearby positions, and then raindrops can splash again reaching other positions.

Initially, a single drop is randomly generated. Then, $N_s$ splashes are randomly generated around it, considering a given maximum splash radius. For each splash whose position is better than the position of the initial drop, the process is repeated; that is, new splashes are created around them. The only differences are that the splash radius is reduced after each step (to intensify the search around promising positions) and that the number $N_s$ of splashes is halved after each step. The pseudocode of SRA can be summarized as follows:

```
initialization of candidate solution
and number of splashes
improving counter = 0
improved solution = False
repeat
  if improved solution then
    decrement improving counter
    generate at most N/2 splashes
    improved = False
  else
    increment improving counter
    generate at most N splashes
  for each splash
    if it improves the current solution then
      move raindrop to splash position
      improved = True
until stopping criteria met
```

This simple algorithm has not attracted much attention from the research community, although it has also been used by other researchers to minimize the risk of distributed denial-of-service attacks [62].

*3.4. WWO: Water Wave Optimization.* The Water Wave Optimization (WWO) metaheuristic was introduced in 2015 in [63]. The method is inspired by the observation of the movement of water waves in both deep water and shallow water. In WWO, it is assumed that the optimization problem is a maximization problem, and the search space is seen as the seabed. In those points where the fitness function is higher, the seabed is also higher, so the amount of water over it is smaller. Thus, promising solutions correspond to shallow water, whereas bad solutions correspond to deep water. We can take advantage of the equations of the movement of water waves to search for the areas with shallow water, i.e., the areas with higher values of the fitness function. In particular, it is known that the propagation of water waves in deep water occurs with larger wavelength and less energy, whereas in shallow water the wavelength is smaller and the energy higher.

Like the WCA metaheuristic, WWO is also a population-based method. All individuals (water waves) are assigned the same initial height and the same initial wavelength. The initial positions of waves are randomly distributed through the search space. In each iteration of the algorithm, three phases occur: propagation, refraction, and breaking. The propagation phase moves the wave to a new position as follows. For each dimension, the position is randomly modified by adding to its previous position a random number between −1 and 1 multiplied by the wavelength and the size of the dimension. Thus, larger wavelengths allow the wave to move farther. Next, the fitness function is evaluated in the new position. In case the solution has improved, the wave is actually moved to the new position; otherwise the wave remains in its original position. When all individuals have gone through the propagation phase, the wavelengths and heights of all of them are updated. The basic idea is that waves moving to better solutions should increase their height and decrease their wavelength. In fact, this operation is done following an equation that takes into account not only the information of each wave, but also the quality of the best and worst individuals, so that a kind of normalization can be done. Ultimately, the equation guarantees that promising waves will propagate within smaller ranges (lower wavelengths) to intensify the exploitation in promising areas, whereas the worst individuals will propagate within larger ranges to improve the exploration of the overall search space.

Regarding the refraction phase, when the wave height decreases to zero, the corresponding wave dies. In this case, the wave is restarted in a new position randomly distributed around the middle point between its previous position and the position of the most promising wave of the population. Its new height is the same as in the initialization phase, while the new wavelength depends on the fitness of both the previous and the current positions: the better the new fitness, the smaller the wavelength, so that better solutions are explored within smaller ranges.

The third phase of each iteration is the breaking phase. In nature, when a wave arrives to a very shallow area, the wave breaks. In WWO, the analogy is that when a wave finds a point with a fitness value better than all previous positions found so far, a breaking phase is performed to intensify the search around such new best position. More precisely, $k$ new waves are created around the new best solution. If none of the new waves outperforms the best solution then they are removed. Otherwise, we have a new best wave. The pseudocode of WWO can be summarized as follows:

```
initialization
compute initial best solution x*
repeat
  for each wave x
    propagate x to a new x'
    if f(x') > f(x) then
      if f(x') > f(x*) then
        break wave x'
        update x*
      replace x with x'
    else
      decrease wave height
      if height == 0 then
```

```
        refract x to a new x'
    update wavelengths
until stopping criteria met
```

WWO has also been used in many different applications, and the original paper [53] has been referenced around 150 times. In addition to classical unimodal and multimodal functions, it has been used for scheduling high-speed trains [63], to classify human activity by analyzing accelerometer information [64], or even to design truss structures [65, 66]. As in the case of WCA, it has also been used in the context of power systems to deal with the economic load dispatch problem [67]. Some modifications of the basic algorithm have also been proposed, including hybridization with the sine cosine algorithm [68] and the introduction of a chaotic phase [69].

*3.5. WEO: Water Evaporation Optimization.* The Water Evaporation Optimization (WEO) was introduced in 2016 in [17]. This metaheuristic is based on the evaporation behavior of a very tiny amount of water deposited on top of a solid surface. Depending on the degree of hydrophobicity/hydrophilicity of the surface, the equation governing the evaporation rate changes in a nonintuitive way. In fact, as the degree of hydrophilicity increases, it would be expected that the evaporation degree is reduced (as the surface tends to keep the water). However, it is known that there is a transition phase, so that, up to a given degree, the evaporation rate increases as hydrophilicity increases. Then, after this transition phase, the water structure becomes monolayer, and the evaporation rate decreases (as expected) as hydrophilicity increases.

Taking into account this behavior, WEO proposes a population-based metaheuristic where individuals are water molecules. In WEO, it is assumed that the optimization problem is a minimization problem. The solid surface where water molecules lie represents the search space: for each position of the search space, the fitness function represents the wettability of such point. That is, the higher the fitness function, the higher the hydrophobicity of that point.

Inspired by the transition phase obtained while changing the wettability of the surface, the overall algorithm is divided into two phases with the same number of iterations in each phase: monolayer evaporation phase and droplet evaporation phase. The first phase represents surfaces with hydrophilicity values higher than the value of the transition phase. Thus, as the hydrophilicity increases, the evaporation rate decreases. The second phase represents surfaces with lower hydrophilicity values. Consequently, as the hydrophilicity decreases, the evaporation rate decreases. In both phases, when a molecule evaporates, it is created again in a different position, although a different method is used for each phase. In the first phase, molecules can be renewed in positions located farther away, whereas in the second phase the renewing method tends toward closer positions. Thus, the method tries to balance local and global search. Besides, in both phases the most promising water molecules are renewed more locally, whereas the worst individuals are renewed farther. The pseudocode of WEO can be summarized as follows:

```
initialization

%monolayer evaporation phase
for i in maxIterations/2
    compute substrate vector normalizing fitness values
    compute evaporation probability matrix
    compute permutation matrix to regenerate molecules
    generate evaporated molecules
    compare and update water molecules

%droplet evaporation phase
for i in maxIterations/2
    compute contact angles normalizing fitness values
    compute droplet evaporation probability matrix
    compute permutation matrix to regenerate molecules
    generate evaporated molecules
    compare and update water molecules
```

Even though the method was proposed only two years ago, the original paper [17] has been referenced more than 40 times by different researchers. In addition to dealing with typical unimodal and multimodal constrained and unconstrained functions (see [17, 70]), it has also been used to solve the truss weight minimization problem where some given stress constraints are to be met [71]. Although most papers using WEO were developed by the original author of the metaheuristic, other researchers have also used it. In particular, the area of power plants appears again as a typical application area of water-based algorithms. In this case, the optimal power flow problem is solved in [72], where several parameters have to be adjusted to optimize the power flow satisfying a given set of constraints. Analogously, in [73, 74] WEO is used to minimize the economical fuel cost and to minimize the emissions produced by a power plant when multiple fuel options are available.

*3.6. RFO: Rainfall Optimization.* The Rainfall Optimization (RFO) metaheuristic was introduced in 2017 in [10]. The method tries to emulate how water flows down from high positions toward lower positions (e.g., going down from mountains to valleys), simulating the tendency to go down through the steeper slopes. In fact, the movement of the water drops is very related to well-known methods like hill climbing or gradient descent. Initially, it rains randomly around all the search space. Then, each drop tries to move to the position in its neighborhood with the lowest value. In order to do that, the neighborhood is defined as all points whose distances to the current position are lower than a given radius $r$, where the value of $r$ is slightly reduced in each iteration, so that the local search is intensified during the last steps of the algorithm.

Let us note that it is not possible to check all points in the neighborhood in a continuous environment. Thus, in each iteration each drop selects randomly $np$ points in its neighborhood. In case the best of those points is better than the current position, the drop moves to it. Otherwise, the method of evaluating points in the neighborhood is repeated up to $Ne$ times, increasing also in each repetition the number $np$ of points to be checked. After $Ne$ repetitions without finding a better position, the drop gets inactive.

After each iteration, drops are sorted according to their ranks. Each rank is computed by taking into account two values: the fitness of the drop position and the difference between its current fitness and its first fitness. Hence, it is not only relevant to the quality of the current solution, because drops that have improved more during their life are considered more promising even if their current fitness is not so high. After creating the merit order list, the lowest ranking drops are turned into the inactive mode, while the individuals with the best rankings are assigned higher values of $Ne$, so that their neighborhoods can be explored in a deeper way.

The algorithm finishes after a given number of iterations or when there are not any active drops in the system.

An interesting concept of RFO is the merit list, where the interest of individuals depends not only on their fitness function (as usual in most metaheuristics) but also on its derivative along time: for individuals which are improving a lot their solutions are considered promising even if their current quality is not good enough. This concept could also be exported to other metaheuristics, in particular to dedicate more resources to those individuals improving their results. The pseudocode of RFO can be summarized as follows:

```
initialization
repeat
  for each active drop
    initialize np value
    repeat
      generate np neighbors
      select best neighbor
      if neighbor improves drop then
        move drop to neighbor
      else
        if Ne iterations reached then
          turn drop inactive
        else
          update np value
    until drop improves or drop inactive
  compute merit list
  turn inactive the worst drops
  increase Ne of the best drops
until all drops are inactive or
maximum iterations reached
```

Even though RFO has been introduced very recently (2017), the first paper describing the method [10] has already been cited 30 times. However, it has not been used much yet. To the best of our knowledge, in addition to the applications presented in the original paper, only one independent research group has used RFO: in [75] RFO is used to solve a variation of the facility location problem.

*3.7. DOA: Droplet Optimization Algorithm.* The last water-based metaheuristic introduced in the literature is the Droplet Optimization Algorithm (DOA), which was published at the end of 2018 in [18]. Although in this case the metaphor is not so clear as in previous metaheuristics, the method gets inspiration from the droplet generation in clouds, the droplet descent from clouds to the ground, and the evaporation.

The algorithm manages two main structures: the list of current drops (candidate solutions) and a list of the best solutions found so far (*BDP*). An initial population of drops is randomly distributed among the search space, and the one obtaining the best fitness among them is included as the only initial element of *BDP*.

In each iteration of the algorithm, the mass center of the current population is computed. Then, the fitness function is evaluated for that point and also for a mutation of that position. The best of both positions is used as the current *big drop* (*BD*). Afterwards, each drop of the population is (partially randomly) moved in the direction toward *BD*. Then, for each candidate drop, two elements of *BDP* are selected, and the position of the candidate drop is modified in a direction that is a combination of its directions toward both elements of *BDP*. In case the new position is worse than its former position, the movement is undone. Otherwise, the process is repeated up to a given number of trials.

The last step of each iteration consists in updating *BDP* to include the best solutions found during the current iteration. If *BDP* reaches a maximum given size then the worse elements of the list are removed. The pseudocode of DOA can be summarized as follows:

```
initialization
introduce in BDP the best drop
repeat
  compute new mass center
```

Table 1: Quantitative comparison of metaheuristics for continuous domain optimization.

| Problem | WCA [15] | SRA [16] | WEO [17] | RFO [10] | DOA [18] |
|---|---|---|---|---|---|
| Sphere | 8,44E-19 | | 6,24E-89 | 1,46E-06 | 1,82E-24 |
| Rosenbrock | 7,00E-06 | | 7,613 | 3,24E+01 | 9,77E-01 |
| Griewank | | 1,44E-02 | 1,11E-17 | 9,86E-02 | 1,77E-02 |
| Rastrigin | 2,00E-07 | 2,62E+02 | 2,65E-01 | | 2,67 |
| Ackley | 1,03E-15 | 1,95E+01 | 8,10E-3 | 1,72E-04 | 2,26E+01 |

```
    compute alternative mass center
    if alternative mass center improves new mass center then
      new mass center = alternative mass center
    for each drop
      move (randomly) toward the new mass center
    for each drop
      trials = 0
      repeat
        choose two big drops from BDP
        compute new position considering such big drops
        if position improves then
          update drop position
          trials ++
      until not improvement or trials >= maximum
    update BDP with the best drops
    update population
  until maximum number of fitness evaluations reached
```

DOA has been proposed very recently. Thus, it has only been used by its creators. In fact, only one paper (see [18]) has been published dealing with DOA, and all the examples presented in it belong to a standard benchmark used at the CEC Conference.

*3.8. Comparison.* In the case of the metaheuristics for continuous domains, the work that the user has to perform to codify solutions is relatively similar in all of them, as opposed to the case of discrete optimization, where the codification of solutions was very different in each metaheuristic. Since conceptual differences among methods are not so radical as in the discrete case (at least, in terms of solution representation), in this case a quantitative comparison is more interesting than a qualitative comparison.

Table 1 shows the results of five of the aforementioned metaheuristics when dealing with five classic optimization functions, all of them minimization problems working on 30 dimensions. The results are taken from the original papers introducing the corresponding metaheuristics. Even though not all the functions were covered by all the metaheuristics, most of them were covered, which allows performing a fair quantitative comparison. Although each result could have been reached after executing the corresponding metaheuristic for a different amount of time (or a different number of fitness function evaluations), we assume the authors did not stop their own algorithms before their results became notoriously stable (i.e., until no significant result improvements were observed for a while), so these values can be fairly taken as the convergence or close-to-convergence results for each metaheuristic and optimization function. Hence, we may fairly compare them in these particular terms (i.e., the speed up to convergence is not considered in the following comparison).

As it can be seen, in most of the cases WCA obtains the best result, while SRA obtains the worst results among all the methods. WEO obtains the best result in the case of the simplest problem (the sphere problem), while in the rest of the cases it obtains satisfactory results, but they are not so good as those of WCA. Finally, RFO and DOA also obtain satisfactory results, but their performance is worse than in the case of WCA.

Regarding the methods not appearing in the table, HCA can also be compared against WCA by using the results appearing in [48]. In this case, HCA obtains nearly the same results as WCA when the number of dimensions of the problem is 2. However, the results of HCA worsen with four dimensions. In fact, HCA has a scalability problem with higher dimensions, so that it cannot deal with the problems of 30 dimensions shown in Table 1. Regarding IWD, there is also a scalability problem, but it is not so relevant as in the case of HCA. In fact, IWD can obtain satisfactory results for up to 10 dimensions, as described in [52].

## 4. Conclusions

In this paper we have introduced all water-based optimization heuristics in the literature we are aware to date, both for combinatorial optimization and for continuous optimization. Most of them are based on emulating, in some way or another, how drops move down due to gravity and form a path on their way down, although other methods simulate other water mechanics. The resulting algorithms are also quite different in general, and basically we may make a distinction between those where the drops (or more general entities) collaboratively form solutions by modifying the values attached to a graph (e.g., in ACO) and those where the drops or other entities are the solutions by themselves, as they define a point in the solution space (e.g., in PSO or hill climbing).

As noted during the previous sections, many algorithms introduce ideas which could be easily exported to other methods with varying degrees of compatibility. In the future, we would like to explore several of these potential hybrid methods, as they could take advantage of the best of different approaches.

## Conflicts of Interest

The authors declare that there are no conflicts of interest regarding the publication of this paper.

## Acknowledgments

This work has been partially supported by the Spanish Ministerio de Economía y Competitividad (project TIN2015-67522-C3-3-R) and by Comunidad de Madrid as part of the program S2018/TCS-4339 (BLOQUES-CM) cofunded by EIE Funds of the European Union.

## References

[1] V. T. Paschos, "An overview on polynomial approximation of NP-hard problems," *Yugoslav Journal of Operations Research*, vol. 19, pp. 3–40, 2009.

[2] D. E. Goldberg, *Genetic algorithms*, Pearson Education, 2006.

[3] M. Dorigo and G. Di Caro, "Ant colony optimization: A new meta-heuristic," in *Proceedings of the 1999 Congress on Evolutionary Computation, CEC 1999*, pp. 1470–1477, USA, July 1999.

[4] M. Dorigo, M. Birattari, and T. Stützle, "Ant colony optimization," *IEEE Computational Intelligence Magazine*, vol. 1, no. 4, pp. 28–39, 2006.

[5] J. Kennedy, "Particle swarm optimization," in *Encyclopedia of Machine Learning*, 766, p. 760, Springer, 2011.

[6] E. W. Dijkstra, "A note on two problems in connexion with graphs," *Numerische Mathematik*, vol. 1, pp. 269–271, 1959.

[7] P. Rabanal, I. Rodrguez, and F. Rubio, "Using river formation dynamics to design heuristic algorithms," in *Proceedings of the International Conference on Unconventional Computation, UC'07*, pp. 163–177, 2007.

[8] H. Shah-Hosseini, "Problem solving by intelligent water drops," in *Proceedings of the Problem solving by intelligent water drops, in: Proceedings of the 2007 IEEE Congress on Evolutionary Computation, CEC'07*, pp. 3226–3231, IEEE, Singapore, September 2007.

[9] F. Yang and Y. Wang, "Water flow-like algorithm for object grouping problems," *Journal of the Chinese Institute of Engineers*, vol. 24, no. 6, pp. 475–488, 2007.

[10] S. H. Aghay Kaboli, J. Selvaraj, and N. Rahim, "Rain-fall optimization algorithm: A population based algorithm for solving constrained optimization problems," *Journal of Computational Science*, vol. 19, pp. 31–42, 2017.

[11] K. Sörensen, "Metaheuristics—the metaphor exposed," *International Transactions in Operational Research*, vol. 22, no. 1, pp. 3–18, 2015.

[12] S. Beucher, C. LantuΘjoul, and C. Lantuéjoul, "Use of watersheds in contour detection," in *Proceedings of the International Workshop on Image Processing, CCETT*, 1979.

[13] L. Vincent and P. Soille, "Watersheds in digital spaces: an efficient algorithm based on immersion simulations," *IEEE Transactions on Pattern Analysis and Machine Intelligence*, vol. 13, no. 6, pp. 583–598, 1991.

[14] I. Kim, D. Jung, and R. Park, "Document image binarization based on topographic analysis using a water flow model," *Pattern Recognition*, vol. 35, no. 1, pp. 265–277, 2002.

[15] A. Sadollah, H. Eskandar, A. Bahreininejad, and J. H. Kim, "Water cycle algorithm with evaporation rate for solving constrained and unconstrained optimization problems," *Applied Soft Computing*, vol. 30, pp. 58–71, 2015.

[16] A. Ibrahim, S. Rahnamayan, and M. V. Martin, "Simulated raindrop algorithm for global optimization," in *27th Canadian Conference on Electrical and Computer Engineering, CCECE14*, pp. 1–8, IEEE, 2014.

[17] A. Kaveh and T. Bakhshpoori, "Water Evaporation Optimization: A novel physically inspired optimization algorithm," *Computers & Structures*, vol. 167, pp. 69–85, 2016.

[18] M. Mohammadpour, H. Parvin, M. Yasrebi, and A. Eskandar Baghban, "Optimisation inspiring from behaviour of raining in nature: droplet optimisation algorithm," *International Journal of Bio-Inspired Computation*, vol. 12, no. 3, p. 152, 2018.

[19] P. Rabanal, I. Rodrguez, and F. Rubio, "Solving dynamic TSP by using river formation dynamics," in *Proceedings of the Solving dynamic TSP by using river formation dynamics, in: Fourth International Conference on Natural Computation (ICNC'08)*, pp. 246–250, IEEE, 2008.

[20] P. Rabanal, I. Rodríguez, and F. Rubio, "Studying the application of ant colony optimization and river formation dynamics to the steiner tree problem," *Evolutionary Intelligence*, vol. 4, no. 1, pp. 51–65, 2011.

[21] H. G. Abood, V. Sreeram, and Y. Mishra, "Optimal placement of PMUs using river formation dynamics (RFD)," in *Proceedings of the 2016 IEEE International Conference on Power System Technology (POWERCON)*, pp. 1–6, Wollongong, NSW, September 2016.

[22] S. Dash, S. Dey, D. Joshi, and G. Trivedi, "Minimizing area of VLSI power distribution networks using river formation dynamics," *Journal of Systems and Information Technology*, vol. 20, no. 4, pp. 417–429, 2018.

[23] S. Dash, D. Joshi, and G. Trivedi, "Multiobjective analog/RF circuit sizing using an improved brain storm optimization algorithm," *Memetic Computing*, vol. 10, no. 4, pp. 423–440, 2018.

[24] S. H. Amin, H. Al-Raweshidy, and R. S. Abbas, "Smart data packet ad hoc routing protocol," *Computer Networks*, vol. 62, pp. 162–181, 2014.

[25] K. Guravaiah and R. Leela Velusamy, "Energy efficient clustering algorithm using rfd based multi-hop communication in wireless sensor networks," *Wireless Personal Communications*, vol. 95, no. 4, pp. 3557–3584, 2017.

[26] P. Rabanal, I. Rodríguez, and F. Rubio, "Testing restorable systems: formal definition and heuristic solution based on river formation dynamics," *Formal Aspects of Computing*, vol. 25, no. 5, pp. 743–768, 2013.

[27] G. Redlarski, M. Dabkowski, and A. Palkowski, "Generating optimal paths in dynamic environments using River Formation Dynamics algorithm," *Journal of Computational Science*, vol. 20, pp. 8–16, 2017.

[28] P. Rabanal, I. Rodríguez, and F. Rubio, "Applications of river formation dynamics," *Journal of Computational Science*, vol. 22, pp. 26–35, 2017.

[29] N. Siddique and H. Adeli, "Water drop algorithms," *International Journal on Artificial Intelligence Tools*, vol. 23, no. 06, Article ID 1430002, 2014.

[30] H. Shah-Hosseini, "The intelligent water drops algorithm: a nature-inspired swarm-based optimization algorithm," *International Journal of Bio-Inspired Computation*, vol. 1, no. 1-2, pp. 71–79, 2009.

[31] S. Niu, S. Ong, and A. Nee, "An improved Intelligent Water Drops algorithm for achieving optimal job-shop scheduling solutions," *International Journal of Production Research*, vol. 50, no. 15, pp. 4192–4205, 2012.

[32] S. Niu, S. Ong, and A. Nee, "An improved intelligent water drops algorithm for solving multi-objective job shop scheduling," *Engineering Applications of Artificial Intelligence*, vol. 26, no. 10, pp. 2431–2442, 2013.

[33] V. Kayvanfar and E. Teymourian, "Hybrid intelligent water drops algorithm to unrelated parallel machines scheduling problem: a just-in-time approach," *International Journal of Production Research*, vol. 52, no. 19, pp. 5857–5879, 2014.

[34] H. Mokhtari, "A nature inspired intelligent water drops evolutionary algorithm for parallel processor scheduling with rejection," *Applied Soft Computing*, vol. 26, pp. 166–179, 2015.

[35] S. R. Rayapudi, "An intelligent water drop algorithm for solving economic load dispatch problem," *International Journal of Electrical and Electronics Engineering*, vol. 5, no. 2, pp. 43–49, 2011.

[36] D. C. Hoang, R. Kumar, and S. K. Panda, "Optimal data aggregation tree in wireless sensor networks based on intelligent water drops algorithm," *IET Wireless Sensor Systems*, vol. 2, no. 3, pp. 282–292, 2012.

[37] S. Salmanpour and H. Motameni, "Optimal path planning for mobile robot using Intelligent Water Drops algorithm," *Journal of Intelligent & Fuzzy Systems: Applications in Engineering and Technology*, vol. 27, no. 3, pp. 1519–1531, 2014.

[38] J. K. Sidhu and H. Kundra, "IWD based image compression," *International Journal of Advances in Computer Science and Communication Engineering*, vol. 1, no. 1, pp. 31–37, 2013.

[39] F. Glover, "Tabu search: A tutorial," *Interfaces*, vol. 20, no. 4, pp. 43–47, 1990.

[40] A. Srour, Z. A. Othman, and A. R. Hamdan, "A water flow-like algorithm for the travelling salesman problem," *Advances in Computer Engineering*, 2014.

[41] T. H. Tran and K. M. Ng, "A water-flow algorithm for flexible flow shop scheduling with intermediate buffers," *Journal of Scheduling*, vol. 14, no. 5, pp. 483–500, 2011.

[42] F. Pargar and M. Zandieh, "Bi-criteria SDST hybrid flow shop scheduling with learning effect of setup times: water flow-like algorithm approach," *International Journal of Production Research*, vol. 50, no. 10, pp. 2609–2623, 2012.

[43] T. Wu, S. Chung, and C. Chang, "A water flow-like algorithm for manufacturing cell formation problems," *European Journal of Operational Research*, vol. 205, no. 2, pp. 346–360, 2010.

[44] C. C. Chang, "A water flow-like algorithm for cell formation, cell layout, and intracellular machine layout problems," *Journal of Convergence Information Technology*, vol. 8, no. 4, pp. 481–489, 2013.

[45] K. M. Ng and T. H. Tran, "A parallel water flow algorithm with local search for solving the quadratic assignment problem," *Journal of Industrial Management Optimization*, vol. 15, no. 1, pp. 235–239, 2019.

[46] M. Zandieh and A. Chensebli, "Reverse logistics network design: a water flow-like algorithm approach," *Opsearch*, vol. 53, no. 4, pp. 667–692, 2016.

[47] C. Kuo and C. Lee, "Network-based type-2 fuzzy system with water flow like algorithm for system identification and signal processing," *Smart Science*, vol. 3, no. 1, pp. 21–34, 2016.

[48] A. Wedyan, J. Whalley, and A. Narayanan, "Hydrological cycle algorithm for continuous optimization problems," *Journal of Optimization*, vol. 2017, Article ID 3828420, 25 pages, 2017.

[49] A. Wedyan, J. Whalley, and A. Narayanan, "Solving the traveling salesman problem using hydrological cycle algorithm," *American Journal of Operations Research*, vol. 08, no. 03, pp. 133–166, 2018.

[50] A. Wedyan, *Hydrological Cycle Algorithm for Solving Optimisation Problems [Ph.D. thesis]*, Auckland University of Technology, 2018.

[51] P. Rabanal, I. Rodríguez, and F. Rubio, "Applying RFD to construct optimal quality-investment trees," *Journal of Universal Computer Science*, vol. 16, no. 14, pp. 1882–1901, 2010.

[52] H. Shah-Hosseini, "An approach to continuous optimization by the intelligent water drops algorithm," *Procedia-Social and Behavioral Sciences*, vol. 32, pp. 224–229, 2012.

[53] H. Eskandar, A. Sadollah, A. Bahreininejad, and M. Hamdi, "Water cycle algorithm–a novel metaheuristic optimization method for solving constrained engineering optimization problems," *Computers & Structures*, vol. 110-111, pp. 151–166, 2012.

[54] D. Karaboga and B. Akay, "A modified Artificial Bee Colony (ABC) algorithm for constrained optimization problems," *Applied Soft Computing*, vol. 11, no. 3, pp. 3021–3031, 2011.

[55] A. Sadollah, H. Eskandar, and J. H. Kim, "Water cycle algorithm for solving constrained multi-objective optimization problems," *Applied Soft Computing*, vol. 27, pp. 279–298, 2015.

[56] O. B. Haddad, M. Moravej, and H. A. Loáiciga, "Application of the water cycle algorithm to the optimal operation of reservoir systems," *Journal of Irrigation and Drainage Engineering*, vol. 141, no. 5, Article ID 04014064, 2014.

[57] A. Khodabakhshian, M. R. Esmaili, and M. Bornapour, "Optimal coordinated design of UPFC and PSS for improving power system performance by using multi-objective water cycle algorithm," *International Journal of Electrical Power & Energy Systems*, vol. 83, pp. 124–133, 2016.

[58] M. A. Elhameed and A. A. El-Fergany, "Water cycle algorithm-based economic dispatcher for sequential and simultaneous objectives including practical constraints," *Applied Soft Computing*, vol. 58, pp. 145–154, 2017.

[59] M. Sarvi and I. N. Avanaki, "An optimized Fuzzy Logic Controller by Water Cycle Algorithm for power management of Stand-alone Hybrid Green Power generation," *Energy Conversion and Management*, vol. 106, pp. 118–126, 2015.

[60] S. Khalilpourazari and M. Mohammadi, "Optimization of closed-loop supply chain network design: a water cycle algorithm approach," in *Proceedings of the 12th International Conference on Industrial Engineering, ICIE 2016*, pp. 41–45, Iran, January 2016.

[61] K. Gao, Y. Zhang, A. Sadollah, A. Lentzakis, and R. Su, "Jaya, harmony search and water cycle algorithms for solving large-scale real-life urban traffic light scheduling problem," *Swarm and Evolutionary Computation*, vol. 37, pp. 58–72, 2017.

[62] S. Bhagat and S. K. Pasupuleti, "Simulated raindrop algorithm to mitigate DDoS attacks in cloud computing," in *Proceedings of the 6th International Conference on Computer and Communication Technology, ICCCT 2015*, pp. 412–418, India, September 2015.

[63] Y.-J. Zheng, "Water wave optimization: A new nature-inspired metaheuristic," *Computers & Operations Research*, vol. 55, pp. 1–11, 2015.

[64] M. Kilany, A. E. Hassanien, and A. Badr, "Accelerometer-based human activity classification using water wave optimization approach in," in *Proceedings of the 11th International Computer Engineering Conference, ICENCO15*, pp. 175–180, 2015.

[65] V. J. Savsani, G. G. Tejani, V. K. Patel, and P. Savsani, "Modified meta-heuristics using random mutation for truss topology optimization with static and dynamic constraints," *Journal of Computational Design and Engineering*, vol. 4, no. 2, pp. 106–130, 2017.

[66] C. Millán Páramo, "Diseño óptimo de armaduras empleando optimización con ondas del agua," *INGE CUC*, vol. 13, no. 2, pp. 102–111, 2017.

[67] M. Siva, R. Balamurugan, and L. Lakshminarasimman, "Water wave optimization algorithm for solving economic dispatch problems with generator constraints," *International Journal of Intelligent Engineering and Systems*, vol. 9, no. 4, pp. 31–40, 2016.

[68] J. Zhang, Y. Zhou, and Q. Luo, "An improved sine cosine water wave optimization algorithm for global optimization," *Journal of Intelligent & Fuzzy Systems: Applications in Engineering and Technology*, vol. 34, no. 4, pp. 2129–2141, 2018.

[69] A. A. Hematabadi and A. A. Foroud, "Optimizing the multi-objective bidding strategy using min–max technique and modified water wave optimization method," *Neural Computing and Applications*, pp. 1–19, 2018.

[70] A. Kaveh, "Water evaporation optimization algorithm," in *Advances in Metaheuristic Algorithms for Optimal Design of Structures*, pp. 489–509, Springer, 2017.

[71] A. Kaveh and T. Bakhshpoori, "An accelerated water evaporation optimization formulation for discrete optimization of skeletal structures," *Computers & Structures*, vol. 177, pp. 218–228, 2016.

[72] A. Saha, P. Das, and A. K. Chakraborty, "Water evaporation algorithm: A new metaheuristic algorithm towards the solution of optimal power flow," *Engineering Science and Technology, an International Journal*, vol. 20, no. 6, pp. 1540–1552, 2017.

[73] R. Venkadesh and R. Anandhakumar, "Economic dispatch with multiple fuel options using water evaporation optimization," *International Journal of Computer Applications*, vol. 165, no. 5, pp. 29–35, 2017.

[74] V. Rajarathinam and A. Radhakrishnan, "Water evaporation algorithm to solve combined economic and emission dispatch problems," *The Global Journal of Pure and Applied Mathematics (GJPAM)*, vol. 13, no. 3, pp. 1049–1067, 2017.

[75] M. Akbari-Jafarabadi, R. Tavakkoli-Moghaddam, M. Mahmoodjanloo, and Y. Rahimi, "A tri-level r-interdiction median model for a facility location problem under imminent attack," *Computers & Industrial Engineering*, vol. 114, pp. 151–165, 2017.

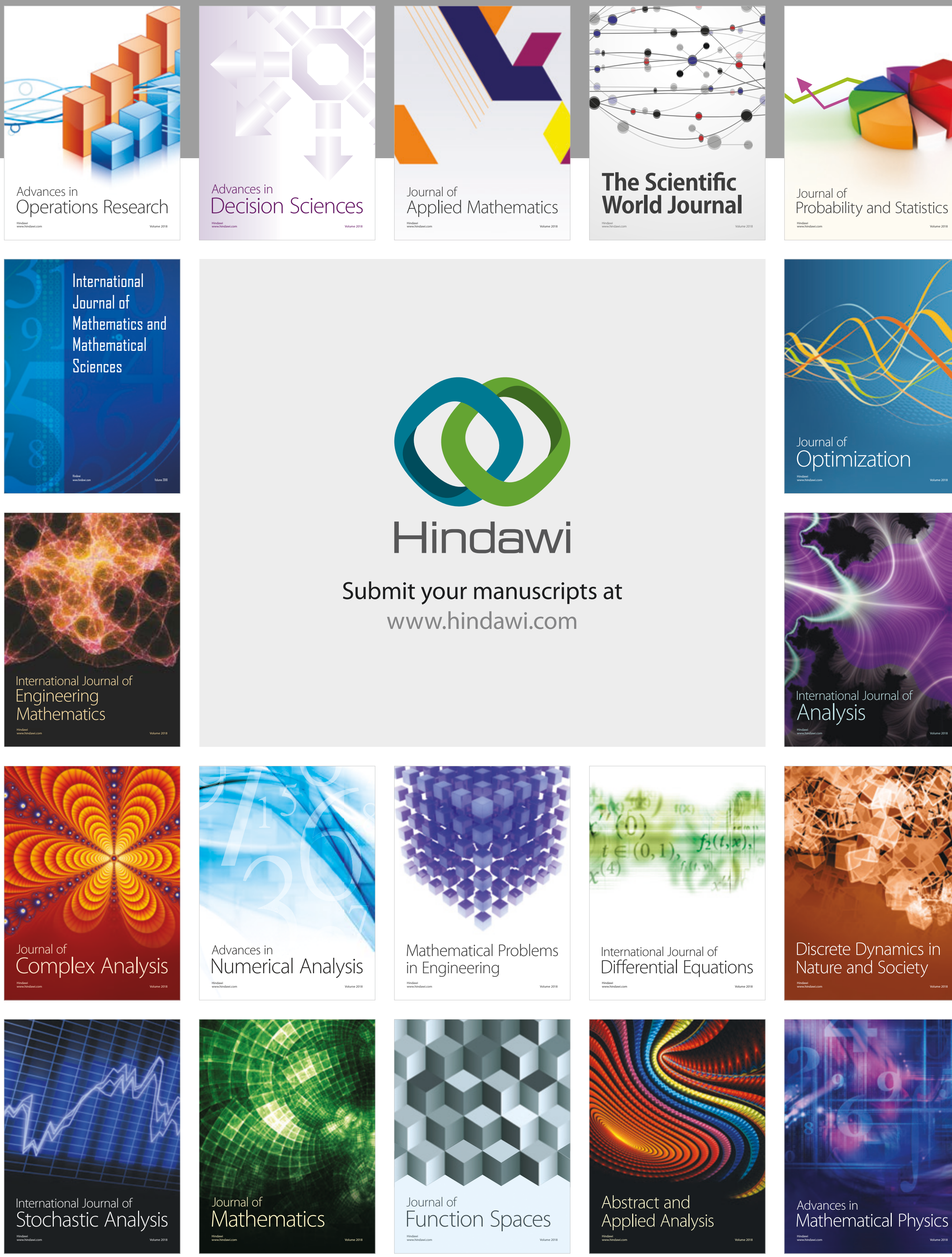
Advances in
Operations Research
Advances in
Decision Sciences
Journal of
Applied Mathematics
The Scientific
World Journal
Journal of
Probability and Statistics
International
Journal of
Mathematics and
Mathematical
Sciences
Hindawi
Submit your manuscripts at
www.hindawi.com
Journal of
Optimization
International Journal of
Engineering
Mathematics
International Journal of
Analysis
Journal of
Complex Analysis
Advances in
Numerical Analysis
Mathematical Problems
in Engineering
International Journal of
Differential Equations
Discrete Dynamics in
Nature and Society
International Journal of
Stochastic Analysis
Journal of
Mathematics
Journal of
Function Spaces
Abstract and
Applied Analysis
Advances in
Mathematical Physics